# B-Splines

Arindam Chaudhuri[1]

[1]Samsung R & D Institute Delhi Noida – 201304 India
`arindamphdthesis@gmail.com`

**Synonyms**

B-Spline 3D curves; B-Spline polygons; B-Spline computer aided design; B-Spline surfaces

**Definition**

B-Splines are one of the most promising curves in computer graphics. They are blessed with some superior geometric properties which make them an ideal candidate for several applications in computer aided design industry. In this article, some basic properties of B-Spline curves are presented. Two significant B-Spline properties viz convex hull property and repeated points' effects are discussed. The B-Splines' computation in computational devices is also illustrated. An industry application based on image processing where B-Spline curve reconstructs the 3D surfaces for CT image datasets of inner organs further highlights the strength of these curves.

**Introduction**

Before the evolution of computer graphics, the aircraft wings and automobile parts were designed through splines. A spline constitutes long wood or plastic pieces of flexible nature where rectangular section is put in place at several positions using heavy lead weights commonly known as ducks. The duck places the spline at fixed positions with respect to the drawing board [1]. This helps spline to take the natural shape considering ducks. The spline's shape can be changed through ducks' movement. This has several drawbacks such as duck positions recording, drafting equipment required towards complex parts, consumer costs, absence of closed form solutions etc [2].

As such polygons give good rendering. But a better way is required towards generating the curved surfaces. For a designer it is difficult to manipulate directly billions of polygons which make up the rendered model. A general way is required to specify arbitrary curved surfaces that can be converted to rendering polygons. For this a mechanism is required which allows to specify any smooth desired curved surface. The solutions are generally provided by three categories of surfaces viz Bézier surfaces, B-

Spline surfaces and subdivision surfaces. In this direction, the computer aided design industry uses NURBS surfaces as its standard definition mechanism. The visual effects industry uses both NURBS and subdivision surfaces. With the introduction of UNISURF which is a CAGD software tool by Pierre Bezier [3], the smooth curves can be easily projected on screens and monitors at low physical storage space. This resulted in the evolution of various CAD based software such as Maya, Blender, 3DMax etc. This followed the development of a new mathematical structure called spline which is a smooth curve represented through few points.

With this motivation, in this article we present the basic properties which define the B-Spline curves. Some of the significant B-Spline properties discussed here are convex hull property and repeated points' effects. The computation task for B-Splines in computational devices is also highlighted. An industry application revolving around image processing with B-Spline applications on 3D surface reconstruction towards inner organs taken from CT images further highlights the significance of these curves. This article is organized as follows. In next section gives an overview of the B-Spline curves with computation task for these curves. This is followed by an industry application involving image processing with B-Splines. Finally, the conclusions are given.

**B-Splines: Overview**

The B-Splines are highly capable for describing various forms of curves [4]. They form a special case for splines, to be more specific the Bezier curves generalization. They are constructed with orthonormal basis of recursive functions. They comprise of curves which are of polynomial nature at points which are basically knots. The polynomial's degree is identical to that of B-Spline. The cubic segments form an important part of B-Splines. Such curves are known as cubic B-Splines. Now we present some B-Splines alongwith certain examples highlighting the special scenarios.

Considering $a, b, c$ coordinates with respect to parameter $t$, B-Spline can be represented as: $a = a(ts), b = b(ts), c = c(ts)$. For B-Spline parametric curve, there exists certain discontinuities for parametric functions at parameter values which depict the knots. For B-Spline, there are several polynomial curve sections as control points number minus polynomial's degree. Considering $N$ control points and $n^{th}$ degree polynomial, there exists $(N - n)$ sections. The joining of these sections happen at $(N - n - 1)$ knots. For control points $\overline{cp}_i; i = 1, \ldots\ldots, N$ with $\overline{cp}_i = \langle x_i, y_i, z_i \rangle$. For polynomial curve's dimensionality connecting knots being $n$, then B-Spline's parametric equation is:

$$\bar{V}(ts) = \sum_{k=1}^{N} BS_{(k-1)}^{n}(ts)\, \overline{cp}_k \quad ts_n \leq ts \leq ts_N, N \geq n+1 \quad (1)$$

The knots should have parameter values $ts$ alongwith them. The $ts$ values at knots are represented as $ts_{j+n}$ considering knot joint of $j^{th}$ and $(j+1)^{th}$ polynomial segments. This is in addition to the parameter values $ts_n$ and $ts_N$ which correspond towards start and end for complete curve. There are also $2n$ parameter values $ts_0, \ldots\ldots, ts_{n-1}$ and $ts_{N+1}, \ldots\ldots, ts_{N+n}$ values which are associated blending polynomials. The values $ts_j$ are monotonically increasing values which may be either equally spaced, integers or positive.

The functions $BS_k^n(ts)$ are recursively defined as:

$$BS_k^n(ts) = \frac{ts - ts_k}{ts_{k+n} - ts_k} BS_k^{n-1}(ts) + \frac{ts_{k+n+1} - ts}{ts_{k+n+1} - ts_{k+1}} BS_{k+1}^{n-1}(ts) \quad (2)$$

With unit step function being defined as:

$$us(ts) = \begin{cases} 1 & ts > 0 \\ 0 & ow \end{cases} \quad (3)$$

The $0^{th}$ order polynomial $BS$ is:

$$BS_k^0(ts) = us(ts - ts_k)\, us(ts_{k+1} - ts) \quad (4)$$

It is to be noted that $BS_k^n \neq 0$ considering range $ts_k < ts < ts_{k+n+1}$. The fig. 1 illustrates some B-Spline curves. The B-Splines are basically characterized by following properties:

(a) They can easily be represented through piecewise polynomial. This allows a B-Spline curve to be represented as a linear combination of number of B-Splines. Any B-Spline curve can be refined as linear combination of piecewise segments as shown in fig. 2. The curves can also be refined through linear operation on control points.

(b) The unit integral can effectively represent a B-Spline as: $\int_{-\infty}^{+\infty} BS_k(ts)\, d(ts) = 1$.

(c) They are non-negative in nature such that: $BS_k(ts) \geq 0$.

(d) B-Splines always partition of unity: $\sum_j BS_k(ts - j) = 1$.

(e) They are highlighted through the support factor: $BS_k(ts) \neq 0, ts \in [0, k]$.

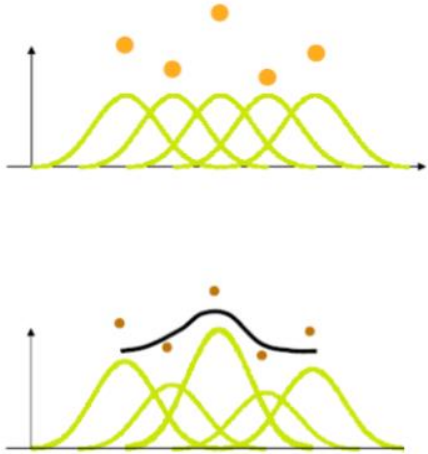

**Fig. 1 Some representative B-Spline curves**

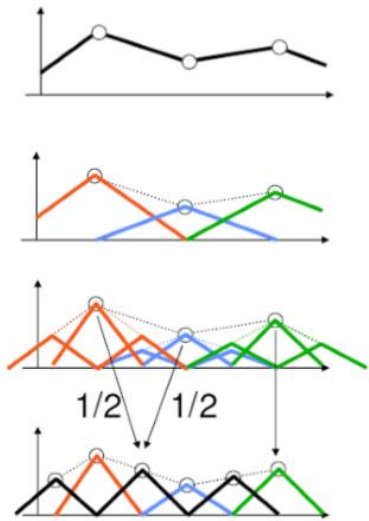

**Fig. 2 B-Spline curve refined as linear combination of piecewise segments**

Now let us discuss some significant properties of B-Splines. The two important properties of B-Splines worth mentioning are the convex hull property and the effect of repeated points. The justifications are trivial in nature.

It is known that by recursion $BS_k^n > 0$ and it can also be shown for one $n$ at a time. As a result of this using recursion leads to:

$$\sum_{k=1}^{N} BS_{(k-1)}^{n}(ts) \equiv 1, \quad ts_n < ts < ts_N \qquad (5)$$

With this B-Splines satisfy the convex hull requirements.

The B-Spline curve moves near the coordinate, when adjacent points of control have identical coordinates. Considering $n$ identical adjacent control points, the spline interpolates the point.

Next, we consider the case where integer knots are equally spaced. Let us consider $ts_k = k$, then form the above equations we have:

$$BS_k^0(ts) = us(ts - k)us(k + 1 - ts) \qquad (6)$$

$$BS_k^n(ts) = \frac{ts-k}{n} BS_k^{n-1}(ts) + \frac{k+n+1-ts}{n} BS_{k+1}^{n-1}(ts) \qquad (7)$$

It is to be noted that $BS_k^0(ts) = BS_0^0(ts - k)$, such that by recursion $BS_k^n(ts) = BS_0^n(ts - k)$. Hence, for knots' integer spacing which results in B-Splines of uniform nature, only single blended polynomial is required towards each spline's degree.

Now some examples are highlighted for cubic B-Splines. The higher degree polynomials become:

$$BS_0^1(ts) = ts\{us(ts)us(1 - ts)\} + (2 - ts)\{us(ts - 1)us(2 - ts)\} \qquad (8)$$

$$BS_0^2(ts) = \frac{1}{2}[ts^2\{us(t)us(1 - ts)\} + \{ts(2 - ts) + (3 - ts)(ts - 1)\}\{us(ts - 1)us(2 - ts)\} + (3 - ts)^2\{us(ts - 2)us(3 - ts)\}] \qquad (9)$$

$$BS_0^3(ts) = \frac{1}{6}[ts^3\{us(ts)us(1 - ts)\} + \{ts^2(2 - ts) + ts(ts - 1)(3 - ts) + (ts - 1)^2(4 - ts)\}\{us(ts - 1)us(2 - ts)\} + \{ts(3 - ts)^2 + (ts - 1)(3 - ts)(4 - ts) + (ts - 2)(4 - ts)^2\}\{us(ts - 2)us(3 - ts)\} + \{(4 - ts)^3 us(ts - 3)us(4 - ts)\}] \qquad (10)$$

The graphical plots corresponding to the four lowest order blending polynomials are shown in the fig. 3.

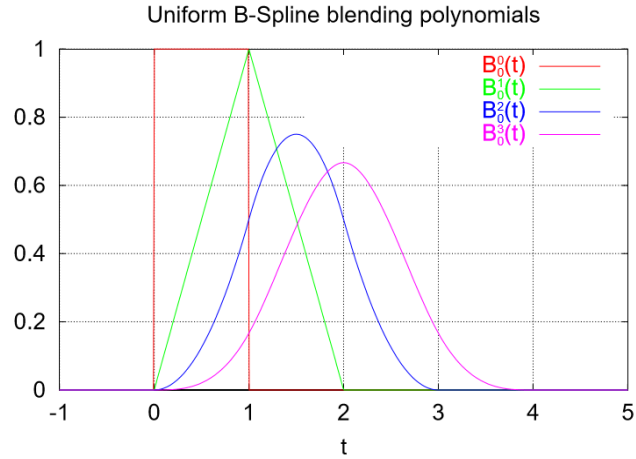

**Fig. 3 The four lowest order blending polynomials corresponding to uniform B-Splines**

With respect to above equations, B-Spline can also be refined as linear combination of dilates and translates of the curve itself. This can be written recursively for $k = 0,1$ as:

$$BS_0(ts) = BS_0(2ts) + BS_0(2ts - 1) \quad (11)$$

$$BS_1(ts) = \frac{1}{2}[BS_1(2ts) + 2BS_1(2ts - 1) + BS_1(2ts - 2)] \quad (12)$$

The above equations are represented through the curves shown in fig. 4. Some of the refinement masks are also shown in fig. 5.

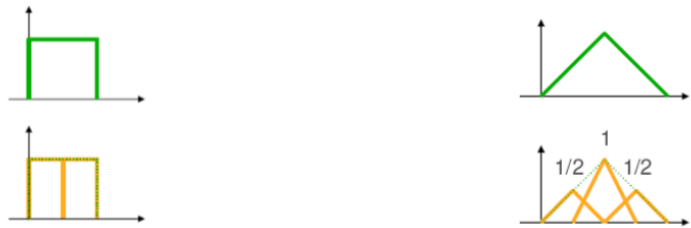

**Fig. 4 The curves on left and right are represented through above equations**

The above equations are further generalized as follows:

$$BS_k(ts) = \frac{1}{2^k}\sum_{k=0}^{n+1}\binom{n+1}{k}BS_k(2ts - k) \quad (13)$$

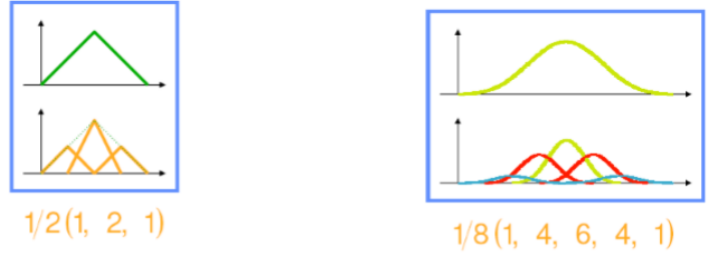

**Fig. 5 The curves on left and right represent the refinement masks for $\frac{1}{2}(1, 2, 1)$ and $\frac{1}{8}(1, 4, 6, 4, 1)$**

The subdivision operators can readily be applied on B-Splines. The subdivision works through bases and control points. Considering a B-Spline, subdivision operator $S$ can be enforced such that:

$$BS(ts) = BS(2ts)S \qquad (14)$$

The term $S$ in above equation represents the subdivision matrix. The fig. 6 represents a typical subdivision matrix. The subdivision operation is stationary in nature.

$$S = \frac{1}{8} \begin{pmatrix} \ddots & \vdots & \vdots & \vdots & \vdots & \ddots \\ \cdots & 1 & 0 & 0 & 0 & \cdots \\ \cdots & 4 & 0 & 0 & 0 & \cdots \\ \cdots & 6 & 1 & 0 & 0 & \cdots \\ \cdots & 4 & 4 & 0 & 0 & \cdots \\ \cdots & 1 & 6 & 1 & 0 & \cdots \\ \cdots & 0 & 4 & 4 & 0 & \cdots \\ \cdots & 0 & 1 & 6 & 1 & \cdots \\ \cdots & 0 & 0 & 4 & 4 & \cdots \\ \cdots & 0 & 0 & 1 & 6 & \cdots \\ \cdots & 0 & 0 & 0 & 4 & \cdots \\ \cdots & 0 & 0 & 0 & 1 & \cdots \\ \ddots & \vdots & \vdots & \vdots & \vdots & \ddots \end{pmatrix}$$

**Fig. 6 A subdivision matrix**

The subdivision can be successively applied on control polygon using control points' sequence. The subdivision in B-Splines leads the resulting curve to convergence, smoothness and approximation. The subdivision matrix converges the curve towards cubic B-spline representation. This allows the control polygon to be drawn instead of the curve. Using an iterative refinement process, the curve approximation is achieved through several refinements as shown in fig. 7. The surface approxima-tion is done using tensor product which is defined over regular quadrilateral meshes. Sometimes semi-regular quadrilateral meshes are used which have few sufficiently separated vertices. The semi-regular meshes represent the boundary surface of an object. In geometric modeling, patches of regular meshes are traditionally used and stuck together. But it becomes difficult towards handling patch density.

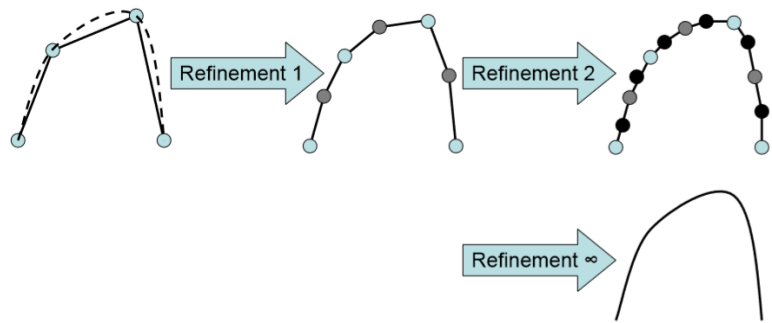

**Fig. 7 The iterative refinement process on curves**

**Computation task for B-splines**

The B-splines are generally computed through fast computing devices [4] because small time steps are required by a smooth curve. To make the computations simple, the calculations are specified with respect to regular uniform B-Splines. Let us start considering the spline order. Considering the order as $k$, a set of $n$ control points are specified. The number of points on the curve may vary depending on the smoothness desired by the user. These set of points are denoted as the curve set.

As the initial step, the uniform knot vector of length $n + k$ is considered. The calculation is performed as follows. Each knot is initialized as zero. Then considering each $1 < i \leq n + k$, it is required to be verified whether conditions $i > n$ and $i < n + 2$ are satisfied. If the conditions hold then current knot takes the value of previous knot with an increment of 1 otherwise, the current knot considers the previous knot value. Now considering $k = n = 4$, the knot vector takes the value $\{0,0,0,0,1,1,1,1\}$. Before calculating the curve set points, it is required to fix a correct step value for parameter $ts$. The step is reached by dividing knot's value considering one less than number of points on curve set.

Now we initiate the computing task towards curve set points for B-spline. Care should be taken such that the number of basis functions which are calculated for each time step is equal that of number of points in control set. Thus considering the entire spline, number of basis computations are same as the product of magnitudes of control and curve sets. Hence for 20 points on curve set with 4 control points, 80 basis functions are required to be calculated.

At each step, Cox-de Boor algorithm is used in order to calculate the basis function's value. During each iteration through the steps, the entire knot vector is taken as with step value $ts$. It is observed that Cox-de Boor algorithm is recursive. As a result of this care needs to be taken such that the basis is calculated correctly. When the set of $n$ basis functions for control points at specified step $ts$ is available, the curve points' coordinates are calculated by multiplying the $i^{th}$ basis function to $i^{th}$ control point. The values obtained are then embedded into curve set. This helps towards generation for collection of points which when plotted together as well as connected with line segments resembles a curve. The number of calculations required to make B-spline are appreciably large.

**Industry Application**

In this section, a novel industry application involving B-Spline surfaces with image processing is presented which is adopted from [5]. The experimental dataset contains about 1900 images from the CT scan of the entire body. Only 500 images considering the chest and upper abdomen have been used for analysis. Here 3D surface is reconstructed for organs developed through tomography. It uses k-means clustering and Hu moments in order to filter contours considering the tomography frames. The least squares B-Spline fitting approximates filtered contours. The organ surface is approximated through B-Spline surface. The technique requires no human intervention. As such high-quality surfaces in versatile format are created.

The contour selection process is a significant step in surface reconstruction. Each contour is selected with utmost care. Then noise is minimized such that resulting surface is not affected. Also, a high-fidelity product needs to be maintained. Here enhancement of each CT slice done and canny edge detector detects the contours. An empirical threshold for edge detector is set. The morphological operations and filtering for small contours is done to eliminate unwanted noise. The effects of filtering are presented in 2[nd] and 3[rd] images for the figs. 8, 9, 10.

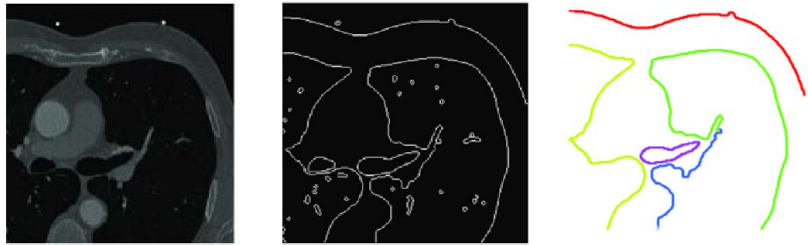

**Fig. 8** Images from 1st dataset (original image, detected edges, detected contours)

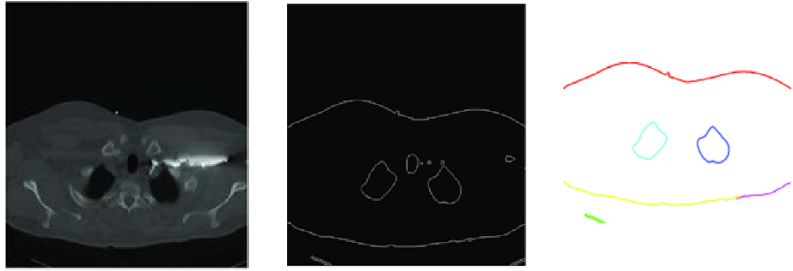

**Fig. 9** Images from 2nd dataset (original image, detected edges, detected contours)

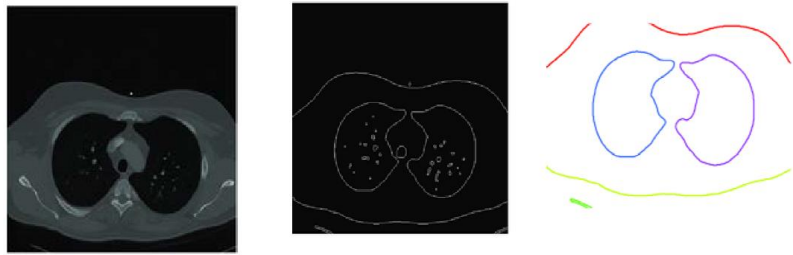

**Fig. 10** Images from 3rd dataset (original image, detected edges, detected contours)

The unwanted contours are removed through automatic noise filtering where larger datasets are created using contour detection. The right lung is taken as region of interest (RoI). About 1996 contours covering chest, left lung, stomach, aorta, etc are created through contour detection and filtering step. The right lung is selected manually from 450 images. Because of shape and contours location diversity, these contours are filtered through k-means clustering. To address the contour's shape, an n-dimen-sional vector is created which is composed of $(X, Y)$ coordinates for contour's centroid and varying Hu moments. After detecting and pre-filtering contours, clustering is done. Considering each contour, Hu moments and centroid are calculated which are used after normalization towards k-means clustering. These clusters are used towards examined contours classification considering euclidean distance from cluster which represents RoI. The testing is done for all combinations of feature vectors based on contour centroid as well as its Hu moments. The best results are achieved through $2^{nd}$ Hu moment with respect to centroid coordinates.

The $n$-dimensional cluster centroids achieved during k-means are utilized towards segmenting the desired contours. Actually, good responses are received with 3D vectors where $2^{nd}$ Hu moment and $(X, Y)$ coordinates for contour's gravity centre are used. The experiments were done using 1996 labeled contours from which 450 are taken as right lung and others are ignored. The figs. 11 and 12 show the correctly classified contours and false +ves classified as contours respectively. It is to be noted that $2^{nd}$ Hu moment alone with $(X, Y)$ coordinates for contour's gravity centre produces appreciable results. The 3D vector substantially reduces the calculations and yields faster and more responsive application.

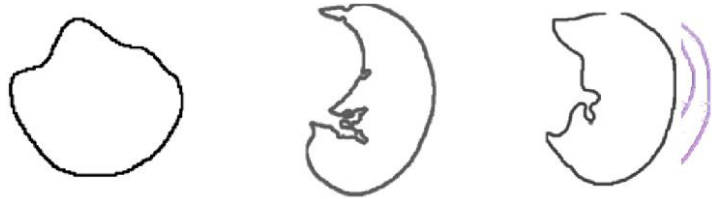

**Fig. 11 The correctly classified contours**

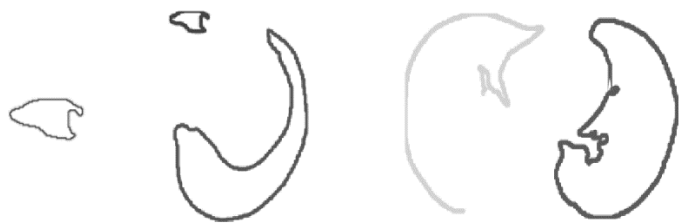

**Fig. 12 The false +ves classified as contours**

After the contour segmentation, the algorithm is approximated and least-squares B-Spline fitting for every contour is done. The contour points in preceeding algorithm steps are used towards fitting cubic B-Spline for every segmented contour cross section.

The curve points form the organ contour are in place and the control points are required to be computed. This can be represented as the following equation where $CV$, $NB$ and $N$ denote the curve, orthonormal basis and control points respectively:

$$CV = NB \cdot N \qquad (15)$$

The above equation can be solved towards $N$ considering least squares such that:

$$N = (NB^T \cdot NB)^{-1} \cdot NB^T \cdot CV \qquad (16)$$

The resulting $N$ matrix comprises of control point vectors towards every cross section. To achieve a smooth surface all approximated cross section splines are made compatible. They are required to have similar degree which is 3 here and need to be specified through identical knot vector. The common knot vector can be achieved by taking the average of the consequent knot vectors considering all the cross sections.

The control points for all cross-section curves are taken column by column and a $2^{nd}$ curves' family is fitted towards approximating these control points columns. This leads to $2^{nd}$ B-Spline curves family. The two control point vectors set forms control net for tensor surface which can be plotted. The control net and 2 knot vectors taken from each curve family completely define the surface. The fig. 13 presents a fragment of right lung surface which is constructed using 3D points. The fig. 14 presents a detailed view of lung fragment with ribs impression in lung surface. The appreciable results are obtained when cross-section contours are approximated through control points with $3^{rd}$ degree B-Spline. The higher order curves look towards greater number of control points and produces curves which are prone with loops and wiggles.

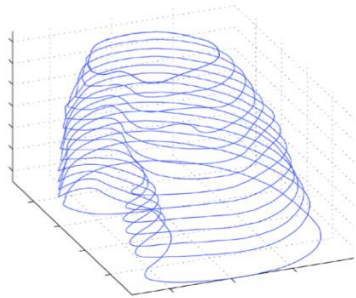

**Fig. 13 The surface approximations through cross-sections**

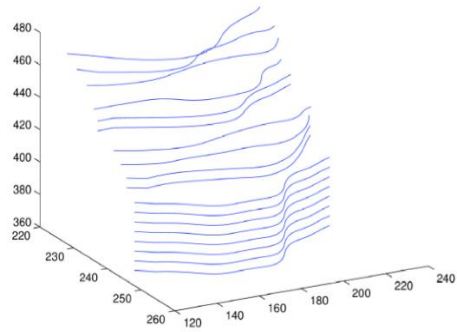

**Fig. 14 The lung surface reconstructed through cross section fragments**

Here there is no need for curve alignment. The fig. 15 represents twisted surface. This does not imply any other problems, except slightly unpleasant visual effects. The fig. 16 shows the lung surface being rendered considering cross sections.

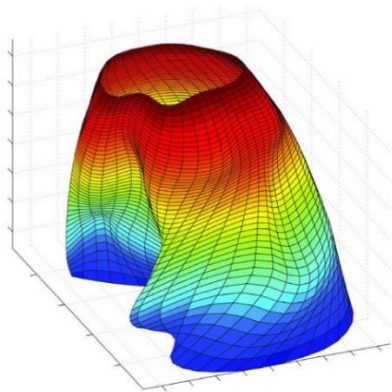

**Fig. 15 The lung's twisted surface**

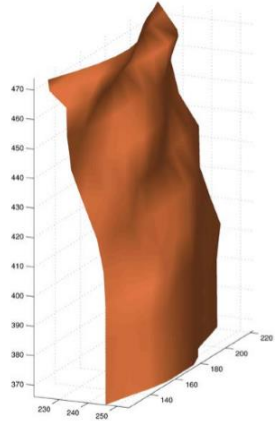

**Fig. 16 The lung surface being rendered considering cross sections**

**Conclusion**

In this article, we have presented some of the important geometric properties of B-Spline curves. The properties worth mentioning are convex hull property and repeated points' effects. These curves find superior applications in computer aided design industry. The computation of B-Splines' in computational devices is also highlighted. An industry application on image processing with B-Spline curve for reconstructing the 3D surface for CT image datasets of inner organs further justifies the importance of these curves.